\documentclass[conference]{IEEEtran}

\IEEEoverridecommandlockouts
\usepackage{cite}
\usepackage{epsfig}
\usepackage{pifont, multirow}
\usepackage{amsmath,amssymb,amsfonts}
\usepackage{algorithmic}
\usepackage{graphicx}
\usepackage{textcomp}
\usepackage{xcolor}
\usepackage{subfig}
\usepackage{array}
\usepackage{tabularx}
\usepackage{hyperref}
\newcommand{\cmark}{\ding{51}}%

\def\BibTeX{{\rm B\kern-.05em{\sc i\kern-.025em b}\kern-.08em
    T\kern-.1667em\lower.7ex\hbox{E}\kern-.125emX}}
\begin{document}

\title{Unified Batch All Triplet Loss for Visible-Infrared Person Re-identification}


\author{\IEEEauthorblockN{Wenkang Li\IEEEauthorrefmark{2}, Ke Qi\IEEEauthorrefmark{2}\textsuperscript{*}, Wenbin Chen\IEEEauthorrefmark{2}, Yicong Zhou\IEEEauthorrefmark{3}}
\IEEEauthorblockA{\IEEEauthorrefmark{2}School of Computer Science and Cyber Engineering\\
Guangzhou University, Guangzhou, China\\
liwenkang25@foxmail.com, qikersa@163.com, cwb2011@gzhu.edu.cn}
\IEEEauthorblockA{\IEEEauthorrefmark{3}Department of Computer and Information Science\\
University of Macau, Taipa, Macau\\
yicongzhou@um.edu.mo}
\thanks{\textsuperscript{*}Corresponding author}
}
\maketitle

\begin{abstract}
Visible-Infrared cross-modality person re-identification (VI-ReID), whose aim is to match person images between visible and infrared modality, is a challenging cross-modality image retrieval task. Batch Hard Triplet loss is widely used in person re-identification tasks, but it does not perform well in the Visible-Infrared person re-identification task. Because it only optimizes the hardest triplet for each anchor image  within the mini-batch, samples in the hardest triplet may all belong to the same modality, which will lead to the imbalance problem of modality optimization. To address this problem, we adopt the batch all triplet selection strategy, which selects all the possible triplets among samples to optimize instead of the hardest triplet. Furthermore, we introduce Unified Batch All Triplet loss and Cosine Softmax loss to collaboratively optimize the cosine distance between image vectors. Similarly, we rewrite the Hetero Center Triplet loss, which is proposed for VI-ReID task, into a batch all form to improve model performance.  Extensive experiments indicate the effectiveness of the proposed methods, which outperform state-of-the-art methods by a wide margin.
\end{abstract}

\section{Introduction}
	Person re-identification is a challenge image retrieval task, whose aim is to match person images across multiple disjoint cameras. These cameras are usually deployed in different locations, so the results of person re-identification can help track the suspects.

	Criminals often collect information during the day and then commit crimes at night, so a reliable day-night person re-identification system is very important. However, ordinary visible light cameras have low visibility at night, so some cameras switch to infrared mode at night. Therefore, the day-night person re-identification task becomes the visible-infrared person re-identification (VI-ReID) task. 

\begin{figure}[t]
\includegraphics[width=8.5cm]{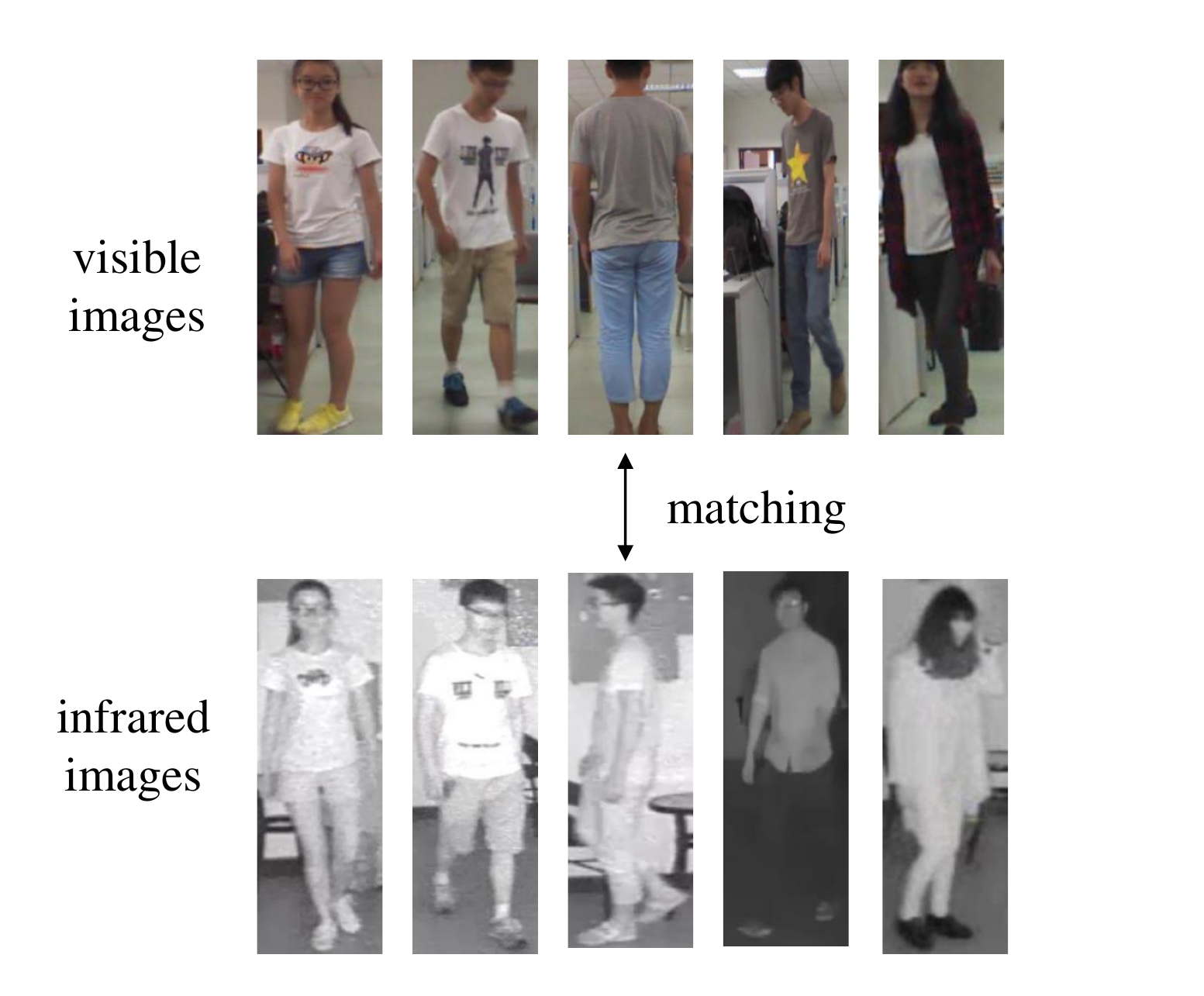}
\caption{Example images of SYSU-MM01~\cite{sysumm01} dataset. Images of visible modality and infrared modality are RGB and grayscale images respectively.}
\label{fig:example}
\end{figure}

	As show in Fig.~\ref{fig:example}, the difference between visible images and infrared images is that infrared images are grayscale images with more noise and less details. To reduce this difference so that the model can learn better, existing works like AlignGAN~\cite{aligngan} translate visible images to infrared images with GAN. However, our experiments show that, in the training phase, simply converting visible images to grayscale images with a certain probability can significantly improve the performance of the VI-ReID model. 

	The basic framework of person re-identification is that, map images into 1D vectors, and then calculate the Euclidean distance or cosine distance between these vectors. The smaller the distance, the higher the similarity between the images. Batch Hard Triplet loss~\cite{triplet} is a loss function that directly optimizes the distance between image vectors, which is widely used in person re-identification tasks. However, we found that it does not work well on the VI-ReID task, because it only optimizes the hardest triplet within a mini-batch, samples in the hardest triplet may all belong to the same modality, which will lead to the imbalance problem of modality optimization. To address this problem, we adopt the batch all triplet selection strategy, which selects all the possible triplets among samples to optimize instead of the hardest triplet within the mini-batch. Similarly, for the VI-ReID task, Hetero Center Triplet loss~\cite{hctri} was proposed, and we also modify it to the batch all version to get better performance.

	In addition to Triplet loss, classification loss such as Softmax loss is also used in most ReID models, which can be regarded as optimizing the distance between each image vector and its class center. But we notice that the metric it optimizing is often different from Triplet loss.  For example, Triplet loss optimizes the Euclidean distance, while the Softmax loss optimizes the vector inner product. In order to unify the metric to be optimized, we introduce Unified Batch All Triplet loss~\cite{circle} and Cosine Softmax loss~\cite{cosinesoftmax} to collaboratively optimize the cosine distance between image vectors.

The main contributions of this paper are summarized as follows:
\begin{itemize}
	\item We find that random grayscale, as a data augmentation method to reduce modality differences, can significantly improve model performance.
	\item We use batch all triplet selection strategy to address the imbalance modality optimization problem of Batch Hard Triplet loss in VI-ReID tasks.
	\item In order to unify the metric to be optimized, we introduce Unified Batch All Triplet loss and Cosine Softmax loss to collaboratively optimize the cosine distance between image vectors.
	\item We rewrite the Hetero Center Triplet loss into a batch all form to improve model performance.

\end{itemize}
 
\section{Related Work}

\subsection{Single-modality Person Re-identification}
The basic framework of person re-identification is that, map images into 1D vectors, and then calculate the Euclidean distance or cosine distance between these vectors. The smaller the distance, the higher the similarity between the images. In order to obtain better features of person, A Omni-Scale feature extraction backbone OSNet~\cite{osnet} was designed. BOT~\cite{bot} uses bag of tricks to build a strong person re-identification baseline model. PCB~\cite{pcb}, MGN~\cite{mgn} and Pyramid~\cite{pyramid} cut the feature maps outputted by backbone into various granularities, and then combine them into the final embedding vector to make better use of local features. AlignedReID~\cite{alignedreid}, PGFA~\cite{pgfa} and CDPM~\cite{cdpm} improve the performance through feature alignment. ABD-Net~\cite{abdnet}, SCAL~\cite{scal}, and SONA~\cite{sona} use attention mechanisms to enhance important areas or channels in the feature map and suppress irrelevant information such as background to obtain more meaningful features. Besides feature representation, loss function is also crucial for person re-identification. Batch Hard Triplet loss~\cite{triplet} and Softmax loss are the two most popular loss functions. The recently proposed Circle loss~\cite{circle} shows good performance on person re-identification.

\subsection{Visible-Infrared Person Re-identification}
In addition to dealing with the common problems of person re-identification, visible-infrared person re-identification also needs to deal with the problems caused by modality differences. Some existing works addressed this by GAN-based methods. AlignGAN~\cite{aligngan} translates visible images to infrared images. D2RL~\cite{d2rl}, Hi-CMD~\cite{hicmd} and JSIA~\cite{jsia} translates visible images and infrared images to each other. cmGAN~\cite{cmgan} only uses adversarial learning to make the features of the two modalities indistinguishable. X Modality~\cite{xmodality}introduces an intermediate modality. Some researchs are about feature learning. EDFL~\cite{edfl} enhances the discriminative feature learning; MAC~\cite{mac} adopts modality-aware collaborative learning; MSR~\cite{msr}learns modality-specific representations. Some other works focus on metric learning. BDTR~\cite{bdtr} calculates the triplet loss of intra-modality and inter-modality respectively; HPILN~\cite{hpiln} calculates the triplet loss of inter-modality in addition to the global triplet loss; HC~\cite{hc} shortens the Euclidean distance between the two modality centers. HC-Tri~\cite{hctri} not only shortens the distance between heterogeneous center of the same class, but also increases the distance between heterogeneous centers of different classes.

\section{Proposed Methods}

\subsection{Unified Batch All Triplet Loss}

\begin{figure}[t]
\centering
\subfloat[Batch Hard]{\label{bh_tri}
\begin{minipage}[c]{.3\columnwidth}
  \centering
  \includegraphics[width=1\textwidth]{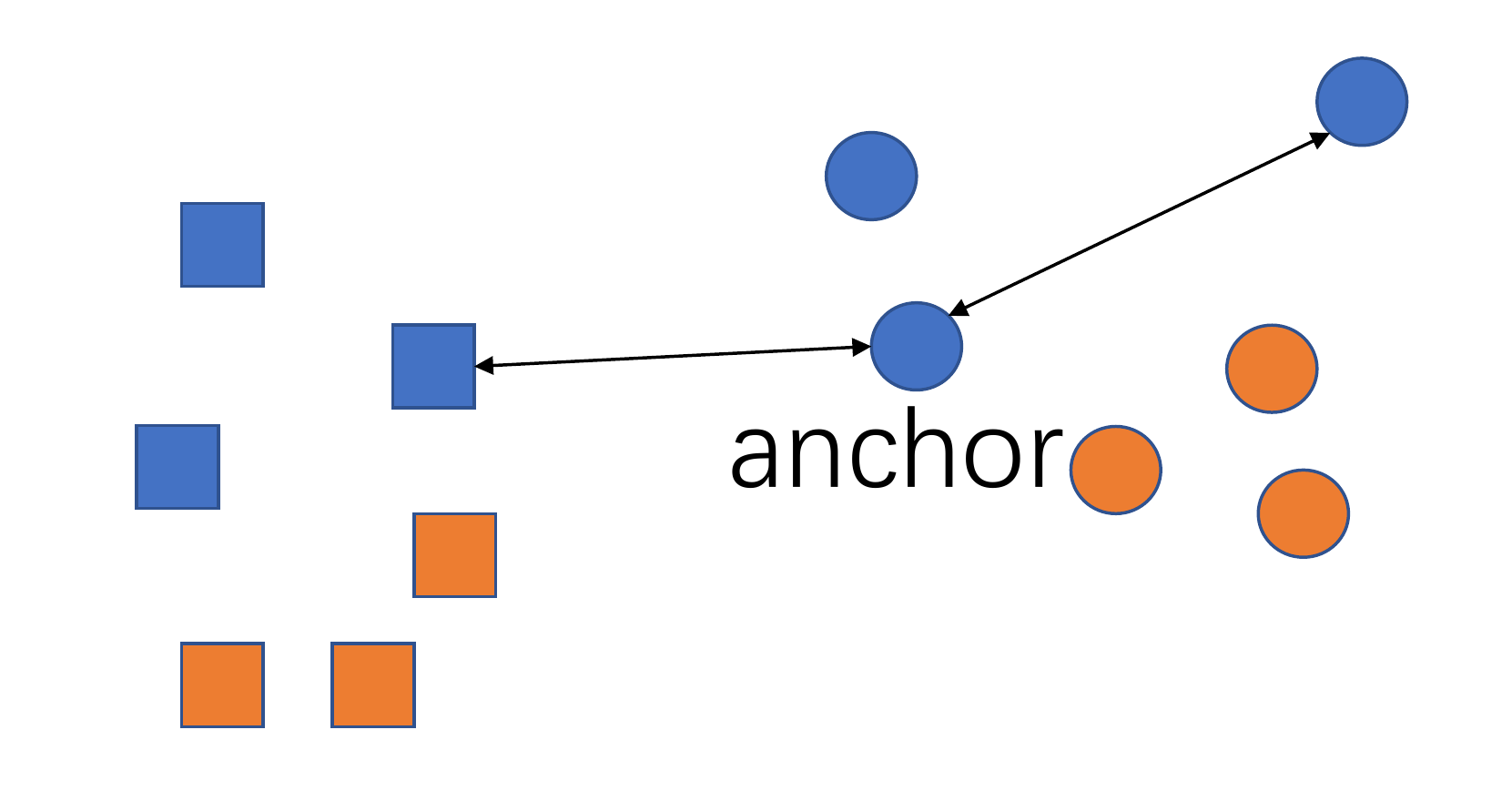}
\end{minipage}
}
\subfloat[CM Batch Hard]{\label{cm_bh_tri}
\begin{minipage}[c]{.3\columnwidth}
  \centering
  \includegraphics[width=1\textwidth]{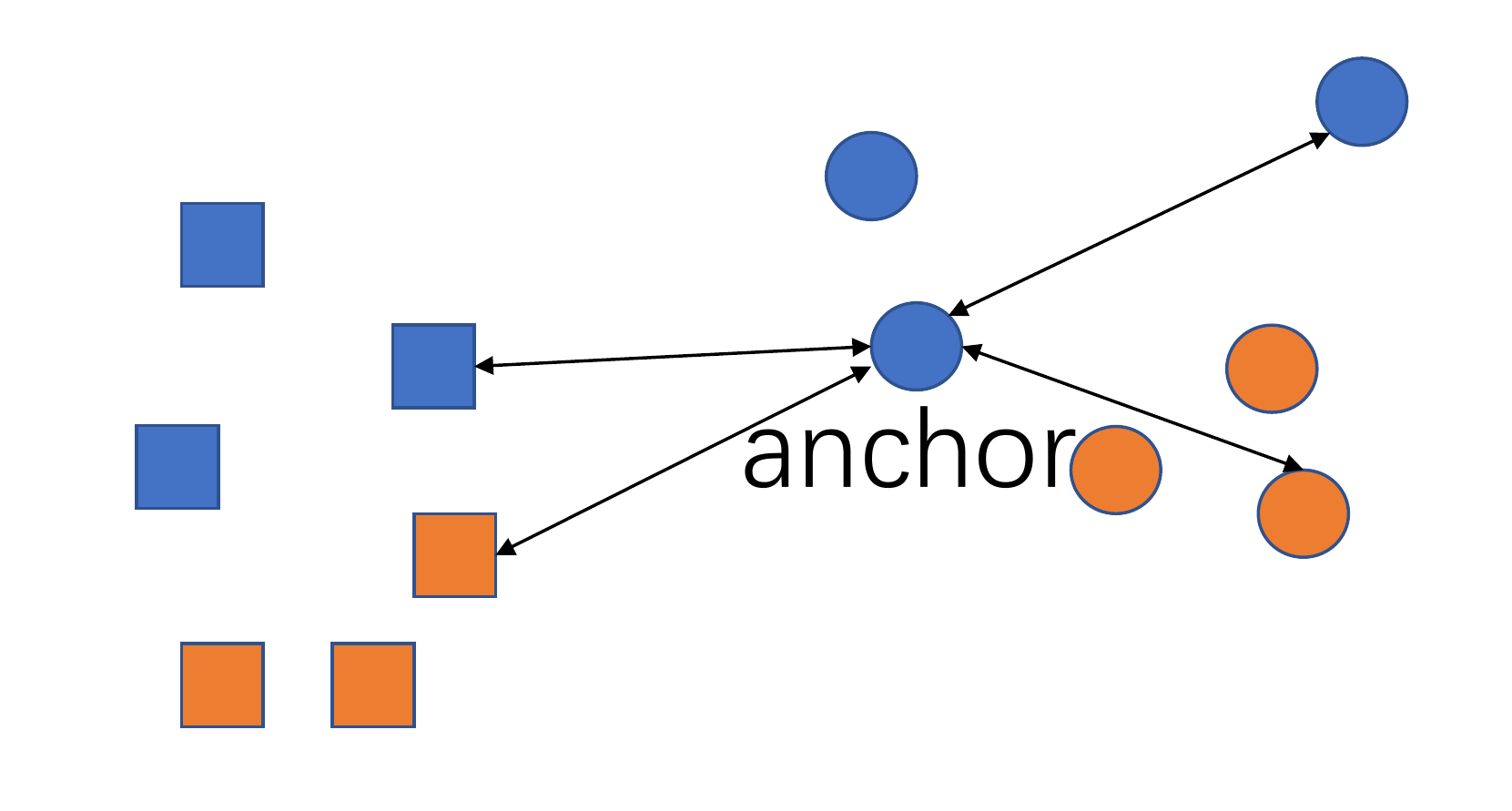}
\end{minipage}
}
\subfloat[Batch All]{\label{ba_tri}
\begin{minipage}[c]{.3\columnwidth}
  \centering
  \includegraphics[width=1\textwidth]{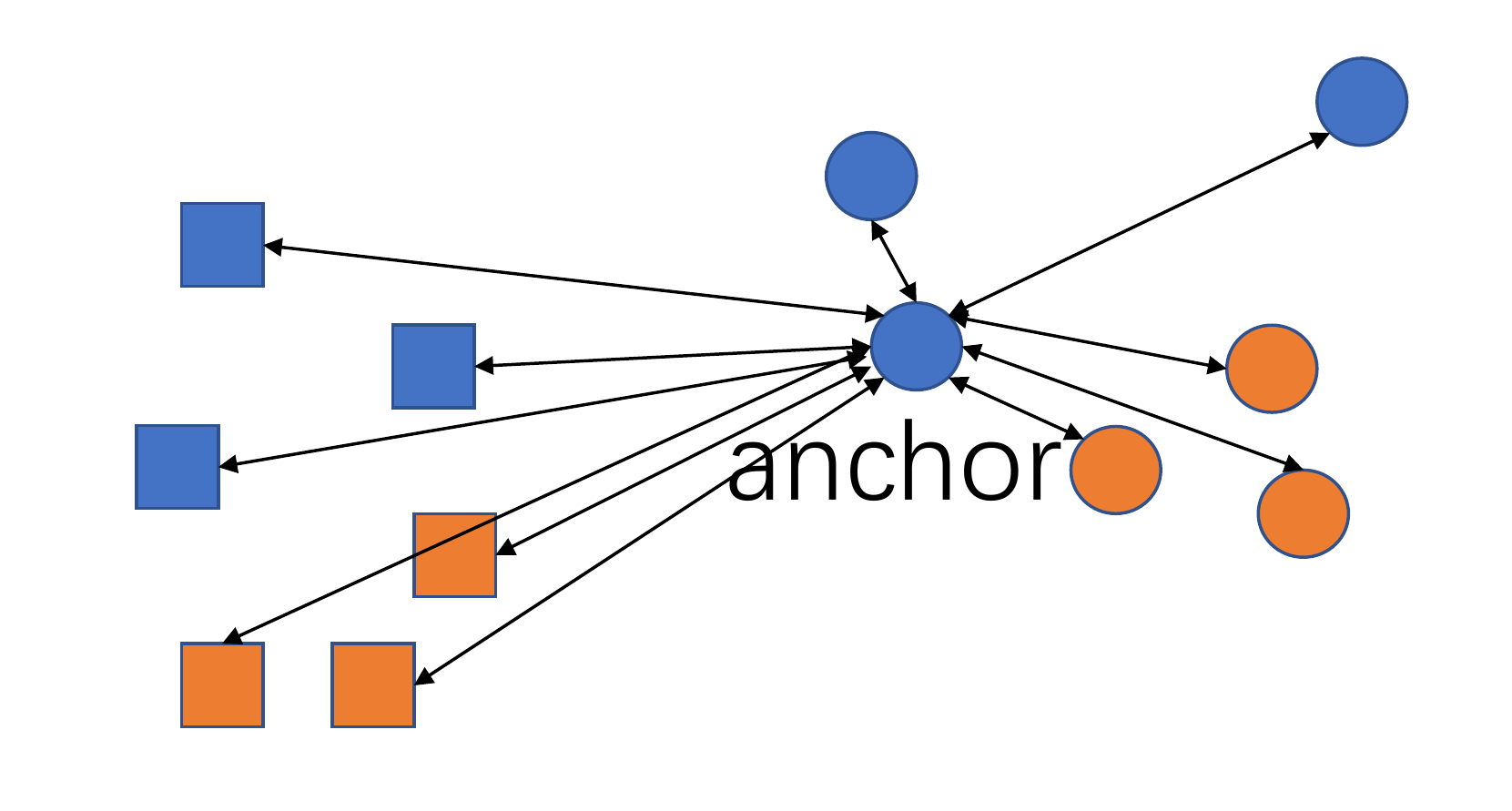}
\end{minipage}
}
\caption{Illustration of different types of triplet selection strategy. Different colors represent different modality, and different shapes represent different IDs. (a) is Batch Hard triplet selection. For each anchor, it select the closest negative sample and the furthest positive sample to form a triplet with anchor. Samples in the selected triplet may all belong to the same modality; (b) is Cross Modality Batch Hard triplet selection. Besides the hardest triplet within the mini-batch, it also select the hardest cross modality triplet for each anchor image. Modality of samples in that triplet is difference with the anchor image; (c) is Batch All triplet selection. It selects all possible triplets among samples.}
\label{fig:triplet}
\end{figure}

Most ReID Models adopt the PK sampling strategy in training phase, which first randomly selects P persons, and then randomly selects K images of each selected person. For VI-ReID tasks, it becomes 2PK sampling strategy, which randomly selects K visible images and K infrared images of each selected person. Such a sampling strategy ensures that each ID has enough and the same number of images in the training phase. Based on this sampling strategy, Batch Hard Triplet loss\cite{triplet} was proposed:
\begin{equation}
\begin{aligned}
L_{bh\_tri} =\overbrace{\sum_{i=1}^{P}\sum_{a=1}^{2K}}^{all\ anchors} \left [m+\overbrace{\max_{p=1...2K}D(x_a^i,x_p^i)}^{furthest\ positive}\right. \\
\phantom{=\;\;}
\left.-\overbrace{\min_{\substack{ n=1...2K \\ j=1...P\\ j\ne i}}D(x_a^i,x_n^j)}^{closest\ negative}   \right ]_+
\end{aligned}
\end{equation}
where $x_a^i$ is the 1D vector of $a^{th}$ image of the $i^{th}$ person in the mini-batch generated by neural network. $D(x, y)$ is the Euclidean distance between vector $x$ and $y$. $\left [x  \right ]_+ =\max(x, 0)$. As shown in Fig.~\ref{bh_tri}, for each anchor image vector, Batch Hard Triplet loss selects the closest negative sample and the furthest positive sample to form the hardest triplet with anchor. If the distance to the furthest positive sample is not smaller than the distance to the closest negative sample by margin term $m$, then penalize these distances. So, Triplet loss pulls the positive sample closer and pushes the negative sample farther to cluster samples. 

The problem of Batch Hard Triplet loss in VI-ReID task is that the samples it selected to form the triplet to optimize may all belong to the same modality. This means that, in each iteration, one of the modalities is optimized more and the other modality is optimized less. Since different modalities follow different distributions and the neural network being used shares the weights between modalities, this kind of optimization imbalance will make the model hard to fit the less optimized modality.

Cross Modality Batch Hard Triplet loss proposed in HPILN~\cite{hpiln} can alleviate this problem. As shown in Fig.~\ref{cm_bh_tri}, besides the hardest triplet within the mini-batch, it also select the hardest cross modality triplet for each anchor image. Modality of samples in that triplet is difference with the anchor image:

\begin{equation}
\begin{aligned}
l_{cm\_bh\_vt}(v_a^i)=\left [ m+ \overbrace{\max_{p=1...K}D(v_a^i,t_p^i)}^{furthest\ infrared\ positive}\right. \\
\phantom{=\;\;}
\left.-\overbrace{\min_{\substack{n=1...K\\j=1...P\\j\ne i} }D(v_a^i,t_n^j)}^{closest\ infrared\ negative} \right ]_+
\end{aligned}
\end{equation}

\begin{equation}
\begin{aligned}
l_{cm\_bh\_tv}(t_a^i)=\left [ m+ \overbrace{\max_{p=1...K}D(t_a^i,v_p^i)}^{furthest\ visible\ positive}\right. \\
\phantom{=\;\;}
\left.-\overbrace{\min_{\substack{n=1...K\\j=1...P\\j\ne i} }D(t_a^i,v_n^j)}^{closest\ visible\ negative} \right ]_+
\end{aligned}
\end{equation}

\begin{equation}
l_{cm\_bh}= \sum_{i=1}^{P}\sum_{a=1}^{K}(\l_{cm\_bh\_vt}(v_a^i) + l_{cm\_bh\_tv}(t_a^i))
\end{equation}

\begin{equation}
L_{cm\_bh}= L_{bh\_tri} + l_{cm\_bh}
\end{equation}
where $v_j^i$ denotes the $j^{th}$ visible image of $i^{th}$ person in the batch, $t_j^i$ denotes the $j^{th}$ infrared image of $i^{th}$ person in the batch. So, $l_{cm\_bh\_vt}(v_a^i)$ selects the hardest infrared triplet for visible image anchor  $v_a^i$ and $l_{cm\_bh\_tv}(t_a^i)$ selects the hardest visible triplet for infrared image anchor $t_a^i$. The final loss $L_{cm\_bh}$ contains the global hardest triplets and the cross modality hardest triplets.

Cross modality batch hard triplet selection can still cause the modality optimization imbalance problem, as long as the global hardest triplet is the same as the cross modality hardest triplet. In this case, Cross modality batch hard triplet selection is downgraded to batch hard triplet selection.

We believe that instead of carefully designing a hard mining strategy, it is better to directly optimize all possible triplets among samples. So, we adopt Batch All Triplet loss~\cite{triplet}:
\begin{equation}
l_{ba\_tri}(x_a^i) = \overbrace{\sum_{\substack{p=1\\p\ne a}}^{2K}}^{all\ pos}\overbrace{\sum_{\substack{j=1\\j\ne i}}^{P}\sum_{n=1}^{2K}}^{all\ neg}\left[m+D(x_a^i, x_p^i) - D(x_a^i, x_n^j) \right]_+ 
\end{equation}

\begin{equation}
L_{ba\_tri} = \frac{1}{2PK} \sum_{i=1}^{P}\sum_{a=1}^{2K}l_{ba\_tri}(x_a^i)
\end{equation}
As shown in Fig.~\ref{ba_tri}, Batch All Triplet loss considers all possible triplets. For each anchor, there are $2K-1$ positive samples and $2(P-1)K$ negative samples, so there are $2(P-1)K(2K-1)$ triplets in total. Because all samples are taken into account, the modality optimization imbalance problem no longer exists.
 
Batch All Triplet loss is fine but there is still a problem, that is, the computational complexity is too high. Each batch needs to calculate a total of $2PK\times 2(P-1)K(2K-1)$ triplets. To solve this problem, we introduce Unified Batch All Triplet loss~\cite{circle}:
\begin{equation}
\begin{aligned}
l_{uni\_ba}(x_a^i)=\overbrace{\sum_{\substack{p=1\\p\ne a}}^{2K}}^{all\ pos}\overbrace{\sum_{\substack{j=1\\j\ne i}}^P\sum_{n=1}^{2K}}^{all\ neg}e^{\gamma(S(x_a^i,x_n^j)-S(x_a^i,x_p^i)+m)}
\end{aligned}
\end{equation}
 \begin{equation}
\begin{aligned}
l_{uni\_ba}(x_a^i)=\overbrace{\sum_{\substack{p=1\\p\ne a}}^{2K}}^{all\ pos}e^{-\gamma S(x_a^i,x_p^i)} \overbrace{\sum_{\substack{j=1\\j\ne i}}^P\sum_{n=1}^{2K}}^{all\ neg}e^{\gamma(S(x_a^i,x_n^j)+m)}
\end{aligned}
\end{equation}

\begin{equation}
\begin{aligned}
L_{uni\_ba}=\frac{1}{2PK}\sum_{i=1}^{P}\sum_{a=1}^{2K}log(1+l_{uni\_ba}(x_a^i))
\end{aligned}
\end{equation}
where $\gamma$ is a scale factor and $m$ is a margin term. $S(x,y)$ computes the cosine similarity between x and y. Similar to Batch All Triplet loss, Unified Batch All Triplet loss also considers all the possible triplets among samples. The difference is that Unified Batch All Triplet loss reduces the computational complexity from $O(2(P-1)K(2K-1))$ to $O(2(P-1)K+(2K-1))$. What's more,  Unified Batch All Triplet loss replaces the hinge function $\left[\bullet\right]_+$ with $exp(\bullet)$, so that all triplets can contribute appropriate loss, rather than triplets larger than margin. It is worth noting that, unlike ordinary triplet loss, which optimizes Euclidean distance, Unified Batch All Triplet loss optimizes cosine similarity. In the next subsection, we will introduce Cosine Softmax loss to collaboratively optimize the cosine similarity. 

\subsection{Collaborative Optimization with Cosine Softmax Loss}

In addition to triplet loss, classification loss such as Softmax loss is also used in most ReID models:
\begin{equation}
L_{sm}=\frac{1}{N}\sum_{i=1}^{N}-log(p_i)=\frac{1}{N}\sum_{i=1}^{N}-log(\frac{e^{f_{y_i}}}{\sum_{j=1}^{C}e^{f_j}})
\end{equation}
where $p_i$ denotes the probability of $x_i$ being correctly classified. N is the number of training samples and C is the number of classes. $y_i$ is the label class of $x_i$. $f_j$ is given by:
\begin{equation}
f_j=W_j^Tx
\end{equation}
where $W\in R^{D\times C}$ is the weights of the fully-connected layer. $W_j$ denotes the $j^{th}$ column in $W$ and $W_j$ can be regarded as the center of class $j$. So Softmax loss optimizes the inner product of each image vector and its class center.

Softmax loss is usually used together with triplet loss. However, Softmax loss optimizes the vector inner product while the triplet loss optimizes the Euclidean distance. They are not consistent. What's worse, in the inference stage, L2 normalized vectors are usually used, which means that whether the vector inner product or Euclidean distance is used as the metric function in the inference stage, it is equivalent to the cosine distance. This is inconsistent with the metric function used during training. To address these problems, we introduce Cosine Softmax loss~\cite{cosinesoftmax} to collaboratively optimize the cosine distance with Unified Batch All Triplet loss. Cosine Softmax loss can be formulated as:

\begin{equation}
L_{cos}=\frac{1}{N}\sum_{i}^{N}-log\frac{e^{\gamma\left(S(W_{y_i}^T,x_i)-m\right)}}{e^{\gamma\left(S(W_{y_i}^T,x_i)-m\right)}+\sum_{\substack{j=1\\j\ne y_i}}^{C}e^{\gamma S(W_j^T,x_i)}}
\end{equation}
where $\gamma$ is a scale factor and $m$ is a margin term. $S(x,y)$ computes the cosine similarity between x and y. The difference between Cosine Softmax loss and Softmax loss is that the former optimizes the cosine distance between the sample vector and the weight vector, while the latter optimizes the vector inner product. In addition, Cosine Softmax loss adds a margin term to make the points belonging to the same class more concentrated.

\subsection{Batch All Hetero Center Triplet Loss}

\begin{figure}[t]
\centering
\subfloat[Batch Hard]{\label{bh_hc}
\begin{minipage}[c]{.48\columnwidth}
  \centering
  \includegraphics[width=1\textwidth]{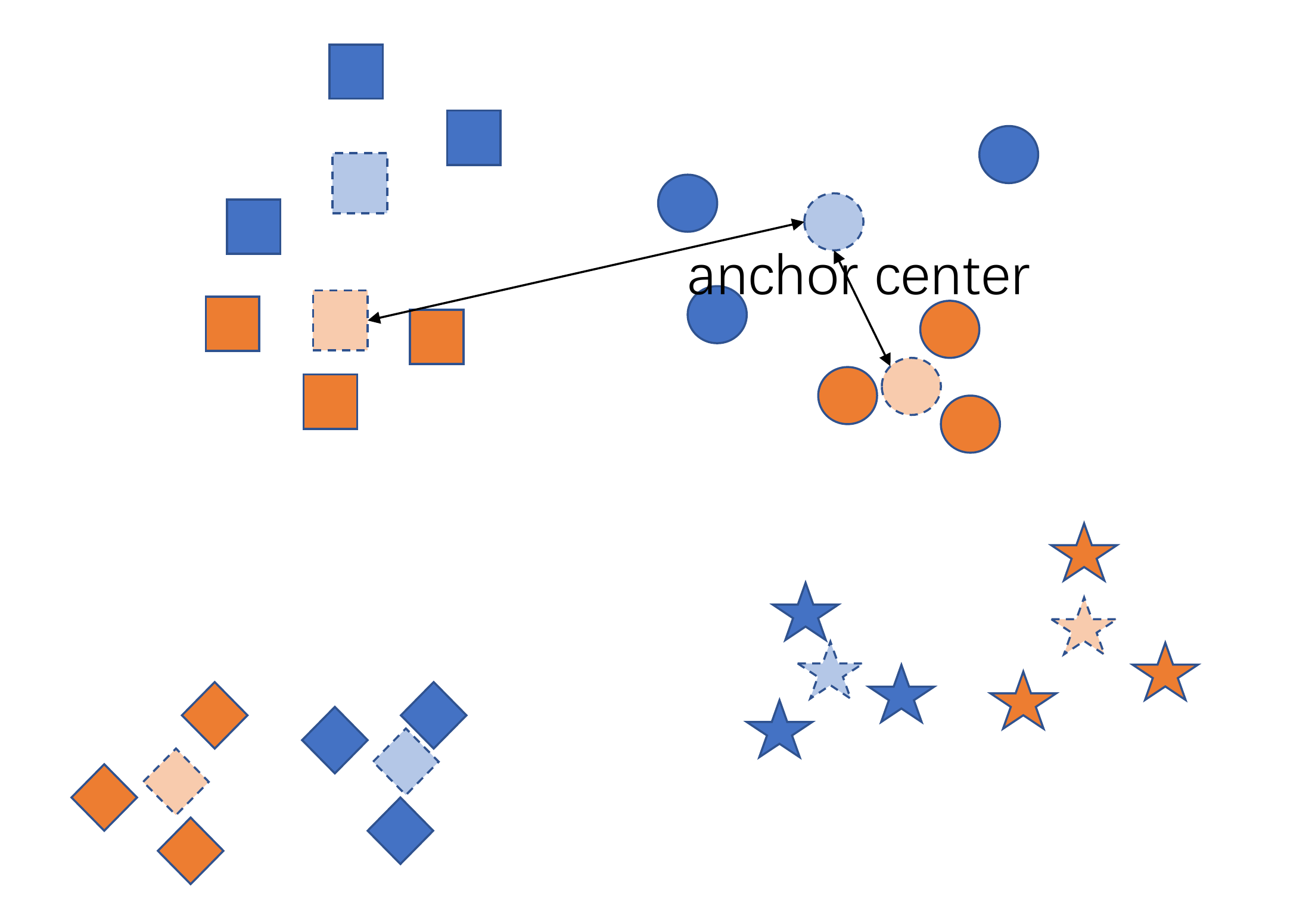}
\end{minipage}
}
\subfloat[Batch All]{\label{ba_hc}
\begin{minipage}[c]{.48\columnwidth}
  \centering
  \includegraphics[width=1\textwidth]{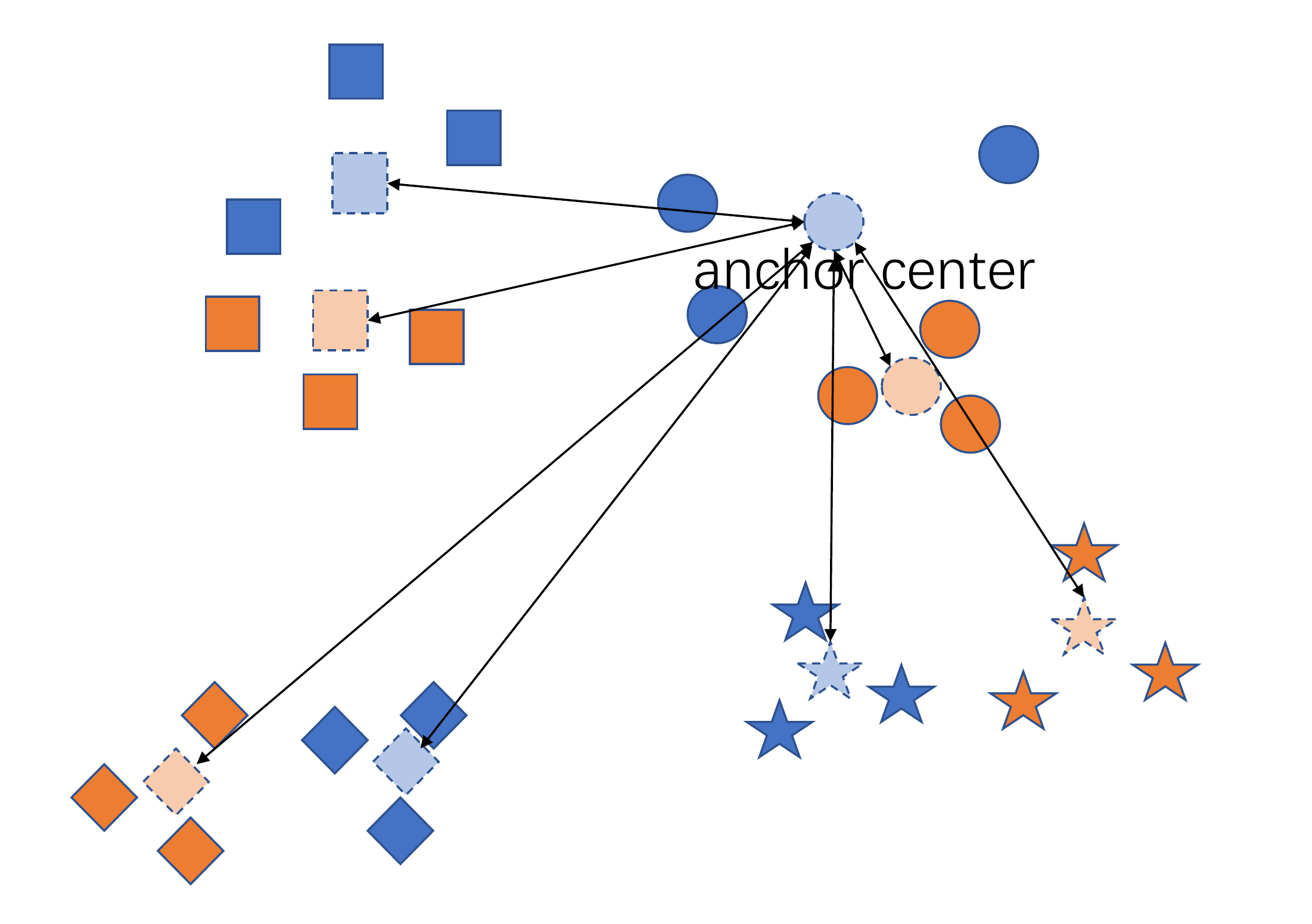}
\end{minipage}
}
\caption{Illustration of different types of center triplet selection strategy. Different colors represent different modality, and different shapes represent different IDs. The light-colored point represents the center of that class in that modality. (a) is batch hard selection. For each anchor center, it selects the center of the same class but in different modality as the positive sample and the hardest center of different classes as the negative sample to form the hardest center triplet. (b) is the batch all selection, it considers all possible center triplets.}
\label{fig:center}
\end{figure}

Vi-ReID task aims to match person images between different modalities, so we do care the distance between samples belonging to the same class but in different modalities. Therefore, Batch Hard Hetero Center Triplet loss~\cite{hctri} was proposed:

\begin{equation}
\begin{aligned}
L_{bh\_hc}=\sum_{i=1}^{P}\left[m+D(c_v^i,c_t^i)-\min_{\substack{n\in\{v,t\}\\j\ne i}}D(c_v^i,c_n^j)\right]_+ \\ + \sum_{i=1}^{P}\left[m+D(c_t^i,c_v^i)-\min_{\substack{n\in\{v,t\}\\j\ne i}}D(c_t^i,c_n^j)\right]_+
\end{aligned}
\end{equation}

\begin{equation}
c_v^i=\frac{1}{K}\sum_{j=1}^{K}v_j^i
\end{equation}

\begin{equation}
c_t^i=\frac{1}{K}\sum_{j=1}^{K}t_j^i
\end{equation}
where $v_j^i$ denotes the $j^{th}$ visible image of $i^{th}$ person in the batch, $t_j^i$ denotes the $j^{th}$ infrared image of $i^{th}$ person in the batch. So, $c_v^i$ is the visible modality center of $i^{th}$ person, $c_t^i$ is the infrared modality center of $i^{th}$ person. $D(x, y)$ is the Euclidean distance between vector $x$ and $y$. As shown in Fig.~\ref{bh_hc}, for each anchor center, Batch Hard Hetero Center Triplet loss selects the center of the same class but in different modality as the positive sample, the hardest center of different classes as the negative sample to form the hardest center triplet. Batch Hard Hetero Center Triplet loss shorten the modality center distance of the same class, increase the center distance between different classes.

\begin{figure*}[t]
\centering
\includegraphics[width=0.8\linewidth]{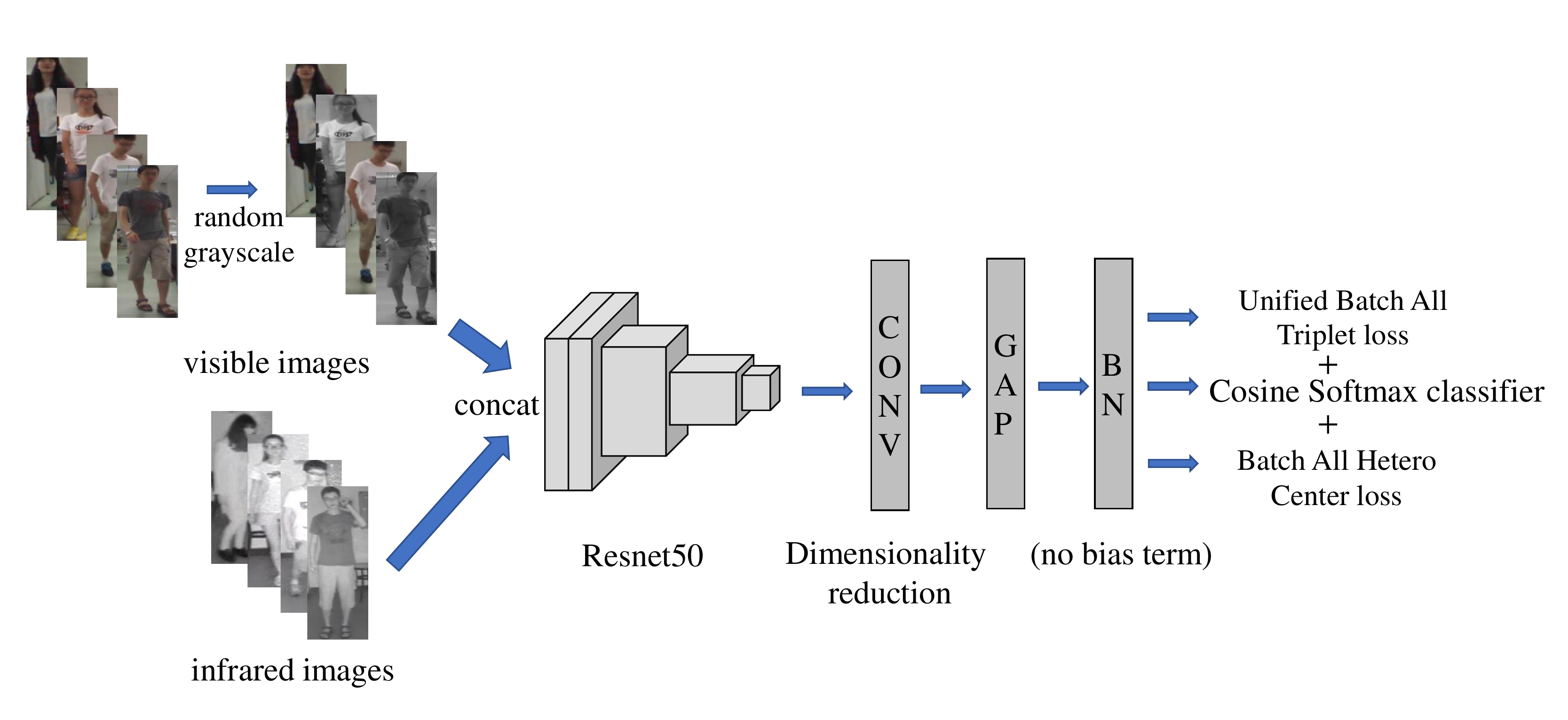}
\caption{Illustration of overall model pipeline. For visible images, we randomly select part of the image with a probability of 0.5 and convert them to grayscale images. then feed all images into the Resnet50 backbone model whose weights are shared by two modalities. For the output feature map, we first use a 1x1 convolutional layer to reduce its dimension to 1024. We use global average pooling to get 1D vectors from the output feature map, and then make it distributed around zero with BN without bias term. Unified Batch All Triplet loss, Batch All Hetero Center loss and Cosine Softmax loss are used together. Note that the Cosine Softmax classifier contains a fully-connected layer.}
\label{fig:pipeline}
\end{figure*}

Similar to triplet loss, we believe that all center triplets should be considered instead of the most difficult center triplet. Therefore, we propose Batch All Hetero Center Triplet loss:
\begin{equation}
\begin{aligned}
L_{ba\_hc}=\sum_{i=1}^{P}log\left(1+\sum_{\substack{n\in\{v,t\}\\j\ne i}}e^{\gamma\left(S(\bar{c}_v^i, \bar{c}_n^j)-S(\bar{c}_v^i, \bar{c}_t^i)+m\right)} \right) \\
+\sum_{i=1}^{P}log\left(1+\sum_{\substack{n\in\{v,t\}\\j\ne i}}e^{\gamma\left(S(\bar{c}_t^i, \bar{c}_n^j)-S(\bar{c}_t^i, \bar{c}_v^i)+m\right)} \right)
\end{aligned}
\end{equation}

\begin{equation}
\bar{c}_v^i=\frac{1}{K}\sum_{j=1}^{K}\bar{v}_j^i
\end{equation}

\begin{equation}
\bar{c}_t^i=\frac{1}{K}\sum_{j=1}^{K}\bar{t}_j^i
\end{equation}
where ${\bar{v}}_j^i$ and $\bar{t}_j^i$ are L2 normalized vectors, so ${\bar{c}}_v^i$ and ${\bar{c}}_t^i$ are centers of L2 normalized vectors. $S(x,y)$ computes the cosine similarity between x and y. Our Batch All Hetero Center Triplet loss has the similar formula as the Unified Batch All Triplet loss. As shown in Fig.~\ref{ba_hc}, it considers all center triplets and optimizes the cosine distance.

\subsection{Model Pipeline}
As shown in Fig.\ref{fig:pipeline}, our model is quite simple and doesn't use advanced methods such as GAN, attention mechanism, local features and multi-branch models. Firstly, we introduce the random grayscale as a data augmentation method to reduce modality differences, which randomly select part of the visible image with a probability of 0.5 and convert them to grayscale images. Then we concatenate infrared images and the processed visible images to form a mini-batch and feed them into the Resnet50 backbone model whose weights are shared by two modalities. Resnet50 outputs a $N\times 2048\times H\times$ feature map. We use a 1x1 convolutional layer to reduce its dimension to 1024. Note that ReLU is used as the activation function and no BN layer after the convolutional layer. We use global average pooling get 1D vectors from the output feature map. Since the last ReLU makes all values of the output not less than 0, we use a BN layer without bias term to make them distributed around zero. Unified Batch All Triplet Loss, Batch All Hetero Center Loss and Cosine Softmax Loss are used in our final model:
\begin{equation}
L = L_{uni\_ba}+L_{cos} + L_{ba\_hc}
\end{equation}
In inference phase, we use the consine distance as the metric function:
\begin{equation}
dist(x, y) = 1 - cos(x, y)
\end{equation}
Including random grayscale, no data augmentation is applied to the test images in inference phase.

\section{Experiments}

\subsection{Experiment settings}

We evaluate our methods on SYSU-MM01~\cite{sysumm01} dataset. The training set of SYSU-MM01 contains 22258 visible images and 11909 infrared images from 395 IDs. The test set contains 6775 visible images and 3803 infrared images from another 96 IDs. We follow the evaluation protocol~\cite{sysumm01} prepared for SYSU-MM01, and mainly report the results of all-search single-shot setting. We report the CMC and mAP metrics.

\subsection{Implement details}

The Resnet50 backbone is initialized with ImageNet pretrained weights. The input images are resized to 320 × 128 for SYSU-MM01. Random erasing and random horizontal flip are adopted as data augmentation. We adopt the 2PK sampling strategy, which first randomly selects P persons, and then randomly selects K visible images and K infrared images of each selected person. We set P=6, K=8 for SYSU-MM01. We use the Adam optimizer with lr=6e-4 and wd=5e-4. We use cosine annealing LR scheduler to train a total of 24 epochs, warm up the first 2 epochs. The margin hyperparameter $m$ is set to 0.3 for all loss. The scale factor $\gamma$ is set to 64 for Cosine Softmax and 12 for Unified Batch All Triplet loss and Batch All Hetero Center Triplet loss.

\subsection{Ablation study}
\subsubsection{Evaluation of each component}

\begin{table}[t]
\begin{center}
{
\caption{Evaluation of each component on the large-scale SYSU-MM01 dataset. We use Batch Hard Triplet loss and Softmax loss as the baseline method. $L_{cos}$ means replacing Softmax loss with Cosine Softmax loss. $L_{uni\_ba}$ means replacing Batch Hard Triplet loss with Unified Batch All Triplet loss. $L_{ba\_hc}$ means adding Batch All Hetero Center Triplet loss. Grayscale means using the random grayscale data augmentation method. Rank1(\%), Rank10(\%) and mAP(\%) Results are reported.} \label{tab:component}
\begin{tabular}{cccc|ccc}
  \hline
   $L_{cos}$& $L_{uni\_ba}$ & $L_{ba\_hc}$ & $Grayscale$ &R1&R10&mAP  \\
	\hline
   & & & & 47.45 & 89.53 & 48.24 \\
	\hline
	\cmark & & & & 56.99 & 89.38 & 53.99 \\  
	 & \cmark& & & 53.25 & 91.48 & 53.93 \\  
	 & &\cmark & & 55.49& 92.56 & 55.32 \\
	 & & &\cmark & 46.83 & 90.58 & 47.25 \\
  \hline
	 \cmark& & &\cmark & 58.28 & 91.14 & 54.76 \\
	 &\cmark & &\cmark & 59.35 & 94.04 & 59.50 \\
	 & & \cmark&\cmark & 56.24 & 93.66 & 55.80 \\
	\hline
	 \cmark& \cmark& & & 59.12 & 91.95 & 58.56 \\
	 \cmark&\cmark & \cmark& & 60.07 & 93.21 & 59.07 \\
	 \cmark&\cmark &\cmark &\cmark & \textbf{65.90} & \textbf{94.52} & \textbf{63.74} \\
	\hline
\end{tabular}
}
\end{center}
\end{table}

Table~\ref{tab:component} shows the effectiveness of each component on the SYSU-MM01 dataset all-search single-shot settings. We use Batch Hard Triplet loss and Softmax loss as the baseline method. $L_{cos}$ means replacing Softmax loss with Cosine Softmax. $L_{uni\_ba}$ means replacing Batch Hard Triplet loss with Unified Batch All Triplet loss.$L_{ba\_hc}$ means adding Batch All Hetero Center Triplet loss. Grayscale means using the random grayscale data augmentation method. We make several observations through the results shown in Table~\ref{tab:component}. 1) Except for the random grayscale data augmentation method, all the proposed methods applied to the baseline model separately can achieve performance improvement. So, all the proposed loss functions are effective. 2) When these advanced loss functions are used together with random grayscale, the performances are better than not using random grayscale.It means that random grayscale can  be effective when it's used with powerful loss functions.  3) When all the proposed methods are used together, we get the best result. It shows that all the proposed methods have no conflict and can be used together.

\subsubsection{Comparison with different triplet losses}
\begin{table}[t]
\begin{center}
{
\caption{Comparison with different triplet losses on SYSU-MM01 dataset. Rank1(\%), Rank10(\%) and mAP(\%) Results are reported.} \label{tab:triplet}
\begin{tabular}{c|c|ccc}
  \hline
   Softmax loss & Triplet loss & R1 & R10 & mAP   \\
  \hline

  \multirow{4}{*}{Softmax}        & Batch Hard & 47.45 & 89.53& 48.24  \\
                                  & CM Batch Hard & 53.19 & 90.63	& 52.20 \\
                                  & Batch All & \textbf{56.61} & 90.50	& \textbf{54.24}  \\
                                   & Unified Batch All & 53.25 & \textbf{91.48}	& 53.93  \\
  \hline
  \multirow{4}{*}{Cosine Softmax}   & Batch Hard & 56.99 & 89.38	& 53.99	 \\
                                    & CM Batch Hard & 56.28 & 89.47	& 53.79	 \\
                                    & Batch All & 57.09 & 89.64	& 54.13	 \\
                                  & Unified Batch All & \textbf{59.12} & \textbf{91.95}	& \textbf{58.56}  \\
  \hline
\end{tabular}
}
\end{center}
\end{table}

Table~\ref{tab:triplet} shows the comparison with different triplet losses on SYSU-MM01 dataset. Random grayscale and Hetero Center Triplet loss are not used. We make several observations through the results. 1) No matter what kind of softmax loss is used, the triplet loss functions using batch all selection strategy are better than the triplet loss functions using batch hard selection strategy. 2) When Unified Batch All Triplet loss is used together with Cosine Softmax loss, the performance is greatly improved. So, it's important  to collaboratively optimize the same distance metric. 3) Except for Unified Batch All Triplet loss, the other triplet losses have similar performance when used with Cosine Softmax loss. It may be because the optimization goals are different, they can not play the expected effect.

\begin{table*}[t]
\footnotesize
\begin{center}
{
\caption{ Comparison with State-of-the-Art Methods on SYSU-MM01 dataset all settings. Rank1(\%), Rank10(\%), Rank20(\%) and mAP(\%) Results are reported. \textsuperscript{*}cmSSFT uses a reranking method while others don’t.} \label{tab:stoa}
\begin{tabularx}{\linewidth}{c|c*{4}{|>{\centering\arraybackslash}X>{\centering\arraybackslash}X>{\centering\arraybackslash}X>{\centering\arraybackslash}X}}
  \hline
	\multirow{3}{*}{Method}&\multirow{3}{*}{Venue}&\multicolumn{8}{c|}{All Search} & \multicolumn{8}{c}{Indoor Search}\\
\cline{3-18}
	& &\multicolumn{4}{c|}{Single-Shot}&\multicolumn{4}{c|}{Multi-Shot}&\multicolumn{4}{c|}{Single-Shot}& \multicolumn{4}{c}{Multi-Shot} \\
\cline{3-18}
	& &R1&R10&R20&mAP&R1&R10&R20&mAP&R1&R10&R20&mAP&R1&R10&R20&mAP\\
\hline
AlignGAN~\cite{aligngan}& ICCV19&42.4&85.0&93.7&40.7&51.5&89.4&95.7&33.9&45.9&87.6&94.4&54.3&57.1&92.7&97.4&45.3\\
Hi-CMD~\cite{hicmd}& CVPR20&34.94&77.58&-&35.94&-&-&-&-&-&-&-&-&-&-&-&-\\
JSIA~\cite{jsia}&AAAI20&38.1&80.7&89.9&36.9&45.1&85.7&93.8&29.5&43.8&86.2&94.2&52.9&52.7&91.1&96.4&42.7\\
XModal~\cite{xmodality}&AAAI20&49.92&89.79&95.96&50.73&-&-&-&-&-&-&-&-&-&-&-&-\\
DDAG~\cite{ddag}&ECCV20&54.75&90.39&95.81&53.02&-&-&-&-&61.02&94.06&98.41&67.98&-&-&-&-\\
HAT~\cite{hat}&TIFS20&55.29&92.14&97.36&53.89&-&-&-&-&62.10&95.75&99.20&69.37&-&-&-&-\\
HC~\cite{hc}&Neuro20&56.96&91.50&96.82&54.95&62.09&93.74&97.85&48.02&59.74&92.07&96.22&64.91&69.76&95.85&98.90&57.81\\
HCTri~\cite{hctri}&TMM20&61.68&93.10&97.17&57.51&-&-&-&-&63.41&91.69&95.28&68.17&-&-&-&-\\
cmSSFT\textsuperscript{*}~\cite{cmssft}&CVPR20&61.6&89.2&93.9&63.2&63.4&91.2&95.7&\textbf{62.0}&70.5&94.9&97.7&72.6&73.0&96.3&99.1&72.4\\
\hline
ours&-&\textbf{65.90}&\textbf{94.52}&\textbf{98.32}&\textbf{63.74}&\textbf{72.42}&\textbf{96.55}&\textbf{99.05}&58.41&\textbf{74.23}&\textbf{98.15}&\textbf{99.62}&\textbf{79.15}&\textbf{83.10}&\textbf{99.09}&\textbf{99.73}&\textbf{73.63}\\
\hline
\end{tabularx}
}
\end{center}
\end{table*}

\subsubsection{Evaluation of Random Grayscale}
\begin{table}[t]
\begin{center}
{
\caption{Evaluation of different grayscale settings on SYSU-MM01 dataset. Training denotes the method used in training phase. Inference denotes what color mode is used by visible images in inference phase. No Grayscale denotes that we don't convert visible images into grayscale images; All Grayscale denotes that all the visible images are converted into grayscale images; Random Grayscale denotes that  we randomly select part of the images with a probability of 0.5 and convert them to grayscale images. Rank1(\%), Rank10(\%) and mAP(\%) Results are reported.} \label{tab:grayscale}
\begin{tabular}{c|c|ccc}
  \hline
   Training & Inference & R1 & R10 & mAP   \\
  \hline

  \multirow{2}{*}{No Grayscale}   & Grayscale & 34.15 & 77.26 & 32.07  \\
                                  & RGB & 59.12 & 91.95	& 58.56 \\
	\hline
  \multirow{2}{*}{All Grayscale}   & Grayscale & \textbf{64.80} & \textbf{94.87} &\textbf{62.62}  \\
                                  & RGB & 57.24 & 93.25	& 54.48 \\
	\hline
  \multirow{2}{*}{Random Grayscale}   & Grayscale &64.39 & 94.44& 61.64  \\
                                  & RGB & 64.78 & 93.85	& 62.46 \\
  \hline
\end{tabular}
}
\end{center}
\end{table}
Table~\ref{tab:grayscale} shows the evaluation of different grayscale settings on SYSU-MM01 dataset. Cosine Softmax loss and Unified Batch All Triplet loss are used in the baseline. Training denotes the method used in training phase. Inference denotes what color mode is used by visible images in inference phase. The following observations can be made: 1) When training with no grayscale, the model biases to the RGB images; When training with all grayscale, the model biases to the gray images; 2) When training with random grayscale, RGB images and grayscale images perform similarly; It means that random grayscale is a strong data augmentation method, which can reduce the modality differences between input images and make the model insensitive to color.

\subsubsection{Comparison with different Hetero Center Triplet losses}

\begin{table}[t]
\begin{center}
{
\caption{Comparison with different Hetero Center Triplet losses on SYSU-MM01 dataset. Cosine Softmax loss, Unified Batch All Triplet loss and random grayscale are used in the baseline. $L_{bh\_hc}$ denotes the Batch Hard Hetero Center Triplet loss and $L_{ba\_hc}$ denotes the Batch All Hetero Center Triplet loss. Rank1(\%), Rank10(\%) and mAP(\%) Results are reported.} \label{tab:center}
\begin{tabularx}{0.9\linewidth}{>{\centering\arraybackslash}X|*{3}{>{\centering\arraybackslash}X}}
  \hline
   Loss Type & R1 & R10 & mAP   \\
  \hline

   $baseline$ & 64.78 & 93.85 & 62.46  \\
   $L_{bh\_hc}$ & 64.74 & \textbf{94.70} & 62.78 \\
\hline
   $L_{ba\_hc}$ & \textbf{65.90} & 94.52 & \textbf{63.74} \\
  \hline
\end{tabularx}
}
\end{center}
\end{table}
Table~\ref{tab:center} shows the comparison with different Hetero Center Triplet losses on SYSU-MM01 dataset. Cosine Softmax loss, Unified Batch All Triplet loss and random grayscale are used in the baseline. $L_{bh\_hc}$ denotes the Batch Hard Hetero Center Triplet loss and $L_{ba\_hc}$ denotes the Batch All Hetero Center Triplet loss. When used with the strong baseline, $L_{bh\_hc}$, which only considers the hardest center triplet, has the best Rank10 performance but gets almost no improvement at Rank1 and mAP, while $L_{ba\_hc}$ , which considers all possible center triplets, has more than 1 point improvement at all metric. It shows the effectiveness of the batch all selection strategy in hetero center triplet loss.

\subsubsection{Comparison with State-of-the-Art Methods}

Table~\ref{tab:stoa} shows the comparison with State-of-the-Art methods on SYSU-MM01 dataset all settings. It shows that, even with a simple network, our proposed method outperforms most State-of-the-Art model at almost all metric except the mAP at all-search multi-shot settings obtained by cmSSFT~\cite{cmssft}. But it is worth noting that cmSSFT includes a reranking technology, which will rank again according to the results of the first ranking. No other methods use reranking. Since our method does not change the network structure, but most other methods change the network structure, this comparison result shows that our method is simple and efficient.

\section{Conclusions}
In this paper, we adopt the batch all triplet selection strategy to solve imbalance modality optimization problem of Batch Hard Triplet loss in VI-ReID tasks. What’s more, in order to unify the metric to be optimized, we introduce Unified Batch All Triplet loss and Cosine Softmax loss to collaboratively optimize the cosine distance. In order to further improve the performance of VI-ReID model, we rewrite the Hetero Center Triplet loss into a batch all form. After using these losses, our model achieves the state-of-the-art results on SYSU-MM01 dataset. We wish that our explorations will benefit the VI-ReID community.

\section*{Acknowledgment}

This work was supported by Science Foundation of Guangdong Province under grant No. 2017A030313374.

\bibliographystyle{IEEEtran}
\bibliography{refer}

\end{document}